\documentclass[journal,twoside,web]{ieeecolor}
\usepackage{generic}
\usepackage{cite}
\usepackage{amsmath,amssymb,amsfonts}
\usepackage{algorithm}
\usepackage{algorithmic}
\usepackage{graphicx}
\usepackage{float}
\usepackage{bm}
\usepackage[font=footnotesize]{caption}
\usepackage{etoolbox}
\AtBeginEnvironment{IEEEeqnarray}{\footnotesize}

\usepackage{setspace}
\usepackage{hyperref}
\hypersetup{hidelinks=true}
\usepackage{textcomp}
\def\BibTeX{{\rm B\kern-.05em{\sc i\kern-.025em b}\kern-.08em
    T\kern-.1667em\lower.7ex\hbox{E}\kern-.125emX}}
\begin{document}
\title{Semantic Alignment of EEG and Text: A Contrastive Learning Framework for Decoding Imagined Speech}
\author{Shahryar G. Namdari*, Francesco Iacomi*, Riccardo Barbieri, \IEEEmembership{Senior Member, IEEE}, and Luca Mainardi, \IEEEmembership{Member, IEEE}
\thanks{* These authors equally contribute to this work.}
\thanks{Shahryar G. Namdari is with the Department of Electronics, Information and Bioengineering, Politecnico di Milano, Milano, 20133, Italy (e-mail: shahryar.namdari@mail.polimi.it).}
\thanks{Francesco Iacomi is with the Department of Electronics, Information and Bioengineering, Politecnico di Milano, Milano, 20133, Italy  (e-mail: francesco.iacomi@polimi.it).}
\thanks{Riccardo Barbieri is with the Department of Electronics, Information and Bioengineering, Politecnico di Milano, Milano, 20133, Italy  (e-mail: riccardo.barbieri@polimi.it).}
\thanks{Luca Mainardi is with the Department of Electronics, Information and Bioengineering, Politecnico di Milano, Milano, 20133, Italy (e-mail: luca.mainardi@polimi.it).}}
\maketitle

\begin{abstract}
Imagined speech decoding from EEG is a promising route toward naturalistic brain--computer interfaces. Traditional closed-set classifiers treat semantic categories as independent labels and therefore do not explicitly exploit the relationships among possible linguistic intents, limiting the flexibility of their output representation.

We adapt a multimodal contrastive learning framework to the imagined-speech setting, aligning EEG trials with language representations by mapping EEG into the embedding space of a frozen pretrained text encoder. An EEG encoder is trained with a composite objective that combines semantically reweighted EEG--sentence alignment with text-prototype-based class discrimination, encouraging semantically structured representations while preserving separability among the 39 intent categories. 

We evaluate the approach on the Chisco imagined speech dataset with 39 semantic categories across multiple subjects, reporting Top-k accuracy, macro-F1, and balanced pairwise matching accuracy (2v2). We additionally examine whether maximum EEG-text cosine similarity provides an intrinsic indicator of decoding reliability and probe semantic structure beyond the original category labels using evaluation-time semantic axes and a qualitative Semantic Gallery Probe (SGP) with previously unseen candidate sentences. The contrastive framework achieved conventional decoding performance comparable to the task-optimized closed-set classifier while mapping EEG into a frozen language embedding space. MaxCosine consistently identified subsets of held-out trials with higher Top-1 and Top-5 accuracy across subjects, evaluation-time semantic axes showed mean agreement above the 50\% chance reference across all five examined dimensions, and the SGP qualitatively illustrated direct comparison with previously unseen candidate sentences. Together, these findings show that EEG--language alignment can retain comparable fixed-intent decoding while enabling similarity-based retrieval, intrinsic reliability ranking, and evaluation of broader semantic structure.
\end{abstract}


\section{Introduction}
\label{sec:introduction}
Brain--Computer Interfaces (BCIs) aim to establish a direct communication channel between the human brain and external devices \cite{wolpaw2002bci}. Beyond motor imagery and P300 paradigms, recent work has increasingly explored decoding of \textit{imagined speech}, the internal articulation of words without vocalization, as a more natural communication pathway, particularly for individuals with severe motor impairments that limit reliable speech output such as Amyotrophic Lateral Sclerosis (ALS) and locked-in syndrome \cite{kubler2001lockedin,chaudhary2016natrevneurol}.

Despite this promise, imagined-speech EEG decoding is still predominantly formulated as a closed-set classification problem, in which semantic categories are treated as independent labels~\cite{lopezbernal2022review}. Such classifiers are effective for fixed-vocabulary intent recognition, but their output representation does not explicitly encode relationships among semantically related intents. A representation aligned with language space offers a complementary objective: preserving discriminability among known categories while organizing EEG trials according to the semantic structure of their linguistic targets.

The recent release of the Chinese Imagined Speech Corpus (Chisco) \cite{zhang2024chisco}, with 39 semantic categories and multiple subjects, enables learning objectives that go beyond rigid classification of a limited set of words. 

In this work, we investigate contrastive EEG--language alignment as an alternative representation for imagined-speech decoding. The proposed framework maps single-trial EEG into the embedding space of a frozen pretrained language model and is trained using a composite objective that combines semantically reweighted EEG--sentence alignment with text-prototype-based class discrimination. This design aims to preserve discriminability among the 39 Chisco intent categories while organizing EEG representations within a structured linguistic space.

The study addresses three complementary questions. First, we test whether semantic alignment can retain conventional decoding performance relative to a task-optimized closed-set classifier, using Top-$k$ accuracy, macro-F1, and balanced pairwise matching accuracy (2v2). Second, we investigate whether the maximum cosine similarity between an EEG embedding and the text-derived category prototypes, already computed during category prediction, can serve as an indicator of prediction reliability and support selective decoding without requiring a separate confidence model. Third, we investigate whether the learned EEG representations exhibit semantic structure beyond the original 39 categories by measuring their agreement with text embeddings along five predefined bipolar semantic dimensions (e.g., practical versus social/emotional), introduced only during evaluation. We also use the Semantic Gallery Probe (SGP) to qualitatively examine their similarity to previously unseen candidate sentences. Together, these analyses evaluate not only whether the intended category can be recovered, but also what additional structure becomes available when imagined-speech EEG is represented directly in language space.

\section{Related Work}
Traditional EEG-based imagined speech decoding has largely emphasized task-specific preprocessing, handcrafted feature extraction, and supervised classification under constrained vocabularies, with substantial variability in protocols and performance across studies \cite{lopezbernal2022review}.

Recent progress in neural language decoding increasingly moves beyond closed-set classification toward \emph{semantic} objectives \cite{camaret2025eegclip, li2026brainmosaic}, where brain activity is mapped into continuous representation spaces learned from large text corpora. In neuroimaging, multiple works have shown that representing words or sentences as vectors in a semantic space enables decoding that generalizes beyond a fixed label set \cite{pereira2018universal,tang2023semantic,ye2025generative}. In parallel, advances in non-invasive EEG-based language decoding have begun to leverage pretrained language models to support open-vocabulary generation or retrieval, particularly in stimulus-driven paradigms such as natural reading on ZuCo-style corpora \cite{zuco2018,zuco2020,duan2023dewave,wang2022openvocab,Chen2025HGRU,zhang2025crossspecies}. These trends motivate representation-level alignment which treats labels as related entities. Recent EEG-language work has also explored semantic intent decoding beyond fixed class vocabularies, using compositional semantic units to reconstruct more flexible linguistic outputs \cite{li2026brainmosaic}.

\subsection{Neural-to-Language Semantic Decoding}
A central theme in modern brain-to-language research is to characterize the representational correspondence between neural activity and the latent spaces of large language models. Early demonstrations in fMRI mapped brain activity to distributional semantic vectors for words and sentences, enabling identification and generalization in a shared semantic space \cite{pereira2018universal}. More recently, non-invasive decoders have reconstructed continuous language from fMRI by linking brain responses to latent semantic representations and then generating text \cite{tang2023semantic}, with subsequent work integrating large language models more directly into the reconstruction stage \cite{ye2025generative}. While originally developed for fMRI, these methodologies provide a foundational template for aligning high-dimensional neural manifolds with linguistic embedded patterns, enabling evaluation based on semantic proximity rather than binary label accuracy.

In EEG, open-vocabulary brain-to-text has been explored, mainly under stimulus-driven protocols (e.g., natural reading on ZuCo \cite{zuco2018,zuco2020}), where EEG segments are aligned to corresponding text tokens or sentences using sequence-to-sequence models with pretrained language backbones \cite{wang2022openvocab,duan2023dewave,Chen2025HGRU}.  While semantic grounding via language models is theoretically sound, its implementation in non-invasive EEG is constrained by the inherently low signal-to-noise ratio and the non-stationarity of the neural signal. Recent imagined speech datasets make it possible to revisit semantic decoding under internally generated language conditions rather than externally driven perception \cite{zhang2024chisco}.
However, most neural-to-language systems either focus on stimulus-driven data or rely on end-to-end generation pipelines, leaving open the question of how reliably semantic structure can be recovered from noisy imagined speech EEG using a shared embedding target.

\subsection{Contrastive Semantic Alignment and Reliability-Aware Evaluation}
Contrastive learning has become a dominant strategy for learning aligned representations from paired but scarce data. This dual-encoder contrastive alignment paradigm has been popularized at scale in multimodal settings such as vision and language (e.g., CLIP~ \cite{radford2021clip} and ALIGN~ \cite{jia2021align}), and has also recently been explored for EEG–language alignment \cite{camaret2025eegclip}. In the modern retrieval-based paradigm, contrastive objectives bridge the modality gap by pulling neural embeddings toward their linguistic counterparts while enforcing separability from semantically distant anchors, facilitating meaning-aware retrieval. This approach has been demonstrated using non-invasive EEG and MEG recordings in word-level decoding by contrasting brain signals against embeddings from a pretrained language model \cite{dascoli2025words}. Related contrastive representation learning has also been investigated for robust EEG decoding in language-adjacent settings such as auditory decoding \cite{bollens2025contrastive}, supporting the view that contrastive alignment provides a robust statistical bridge, grounding noisy, transient brain activity in the structured, persistent semantic manifolds of pretrained models.

Evaluation in semantic decoding often goes beyond Top-$k$ accuracy, using pairwise or retrieval-based tests and statistical validation (e.g., permutation testing) to quantify whether neural representations align with semantics beyond chance \cite{pereira2018universal}. Representational similarity analysis (RSA) offers a standard approach for comparing representational structure via similarity patterns \cite{kriegeskorte2008rsa}. Separately, BCI-oriented evaluation increasingly emphasizes operational metrics that capture realistic use such as utility-based metrics \cite{DalSeno2010Utility}, or including abstention or no-control behavior, rather than accuracy alone \cite{Ma2023BCIUtility}. Despite these advances, traditional imagined-speech research often neglects uncertainty quantification, failing to treat reliability as a primary objective alongside representational alignment.

Together, these studies motivate language-aligned neural representations, but several questions remain open for imagined-speech EEG. First, it is unclear whether mapping EEG into a pretrained semantic space can preserve conventional decoding performance in a realistic multi-class setting when compared with a task-optimized closed-set classifier. Second, the reliability information carried by the resulting similarity scores has received comparatively little attention. Third, it remains unclear whether such representations preserve semantic structure beyond the category labels used during training. We address these questions jointly by comparing closed-set and language-aligned decoding, evaluating MaxCosine as an intrinsic reliability-ranking signal, and probing the learned EEG representations using evaluation-time semantic axes and an unseen semantic gallery.

\section{Methodology}
In this section, we describe the proposed semantic alignment framework for mapping imagined-speech EEG into a frozen language embedding space, the closed-set classifier used for comparison, and the common evaluation protocol.

\subsection{Dataset and Signal Representation}
\label{subsec:dataset}

\begin{figure}[H]
\centerline{\includegraphics[width=\columnwidth]{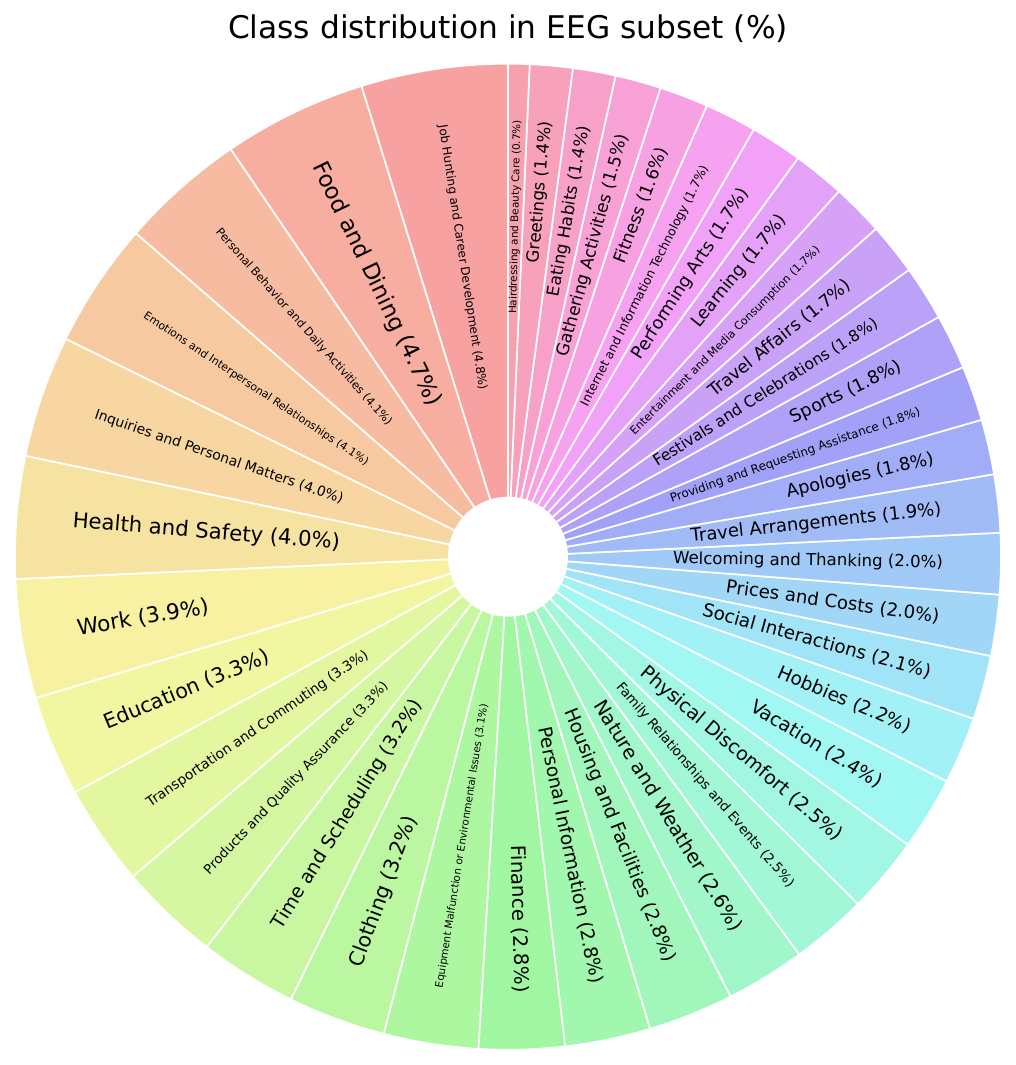}}
\caption[Class distribution across semantic categories]{Semantic classes and their distribution for a representative participant (Subject 1), which is almost consistent across all subjects.}
\label{fig:class_dist}
\end{figure}

This work utilizes the \textit{Chinese Imagined Speech Corpus} (Chisco) \cite{zhang2024chisco}, a comprehensive multi-modal resource developed for imagined speech decoding. The dataset combines synchronized neural recordings with sentences, facilitating systematic analysis of internally generated speech. The \textit{Corpus} contains sentences belonging to 39 semantic categories,  covering diverse conversational contexts (such as Health and Safety, Education, Food and Dining, etc.). As shown in Figure~\ref{fig:class_dist}, the category's distribution exhibits a natural imbalance, with common concepts appearing significantly more frequently than specific situations.

We used version 1.1.2 of the publicly available Chisco dataset (OpenNeuro: ds005170), which contains EEG recordings from five participants. This expanded release includes two additional participants (Sub-04 and Sub-05) beyond the three described in the original dataset publication. All subjects were native Chinese speakers, right-handed, and had normal or corrected-to-normal vision. Prior to data collection, participants underwent a five-day \textit{Integrative Body-Mind Training} (IBMT) program to enhance concentration. The acquisition paradigm follows a cue-based imagined speech protocol: each trial begins with a fixation period to stabilize attention, followed by the presentation of a visual linguistic stimulus (belonging to a specific semantic category). Participants are then instructed to silently and repeatedly read and imagine the displayed sentence without overt articulation or movement. During the trial, EEG signals were acquired using a Compumedics Neuroscan SynAmps-2 system with a 128-channel Quik-Cap (extended 10-20 montage) at a sampling rate of 1,000~Hz.

While the dataset provides data for both the Reading and Imagined Speech phases, in this work we exclusively focused on the imagined speech epochs, corresponding to the 3.3~s window where subjects mentally pronounced the stimuli. We used the  officially released \textit{preprocessed signals}. As detailed in the original documentation \cite{zhang2024chisco}, the raw recordings underwent a standardized cleaning pipeline consisting of channel selection (125 retained channels), downsampling to 500~Hz, bad-channel detection and interpolation with re-referencing using the PREP pipeline \cite{prep2015}, power-line noise removal via a 50~Hz notch filter, high-pass filtering at 1~Hz, segmentation according to the experimental paradigm, automated rejection of corrupted data spans using the Autoreject algorithm \cite{autoreject2017}, and Independent Component Analysis (ICA) with ICLabel-assisted component \cite{iclabel2019} classification and manual correction to remove ocular and muscular artifacts.

Pre-segmented trials were further formatted through channel pruning and normalization. The last three channels were removed from the original 125-channel configuration (vertical electrooculogram (VEO), horizontal electrooculogram (HEO), and Trigger), yielding 122 EEG channels for analysis. Additionally, instance-wise Z-score normalization was applied to each trial to promote consistent input distributions across sessions and subjects.

\subsection{Proposed Contrastive EEG--Language Framework}

\begin{figure}[H]
\centerline{\includegraphics[width=\columnwidth]{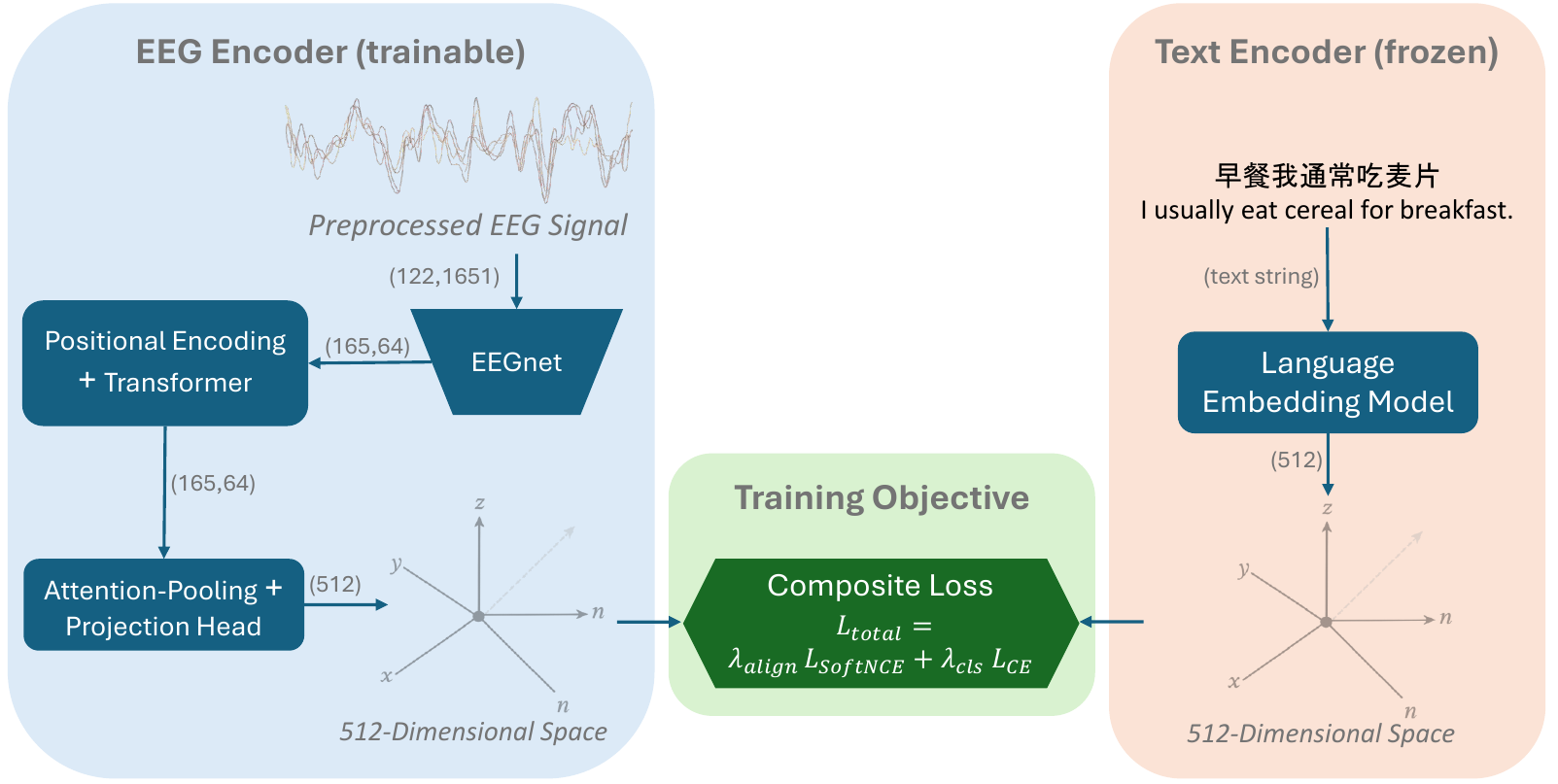}}
\caption[Overview of the semantic alignment framework]{Overview of the proposed Semantic Alignment Framework.}
\label{fig:framework}
\end{figure}

Figure~\ref{fig:framework} provides an overview of the training pipeline for the proposed contrastive framework. The framework receives preprocessed EEG signals as input (left side of the figure) and learns to project them into a continuous semantic space defined by a frozen language embedding model (right part of the figure). Training is driven by a composite objective that encourages alignment between EEG projections and text representations while preserving class-level separability. The following subsections describe the text encoder, EEG encoder, and training objective of the proposed framework in detail.

\subsubsection{Text Encoder}

The target semantic space toward which all EEG representations are aligned is defined by a pretrained language embedding model, denoted as $E_{\text{text}}$. For any stimulus sentence $S$, the encoder produces a fixed-size embedding vector:
\begin{equation}
\mathbf{t} = E_{\text{text}}(S),
\end{equation}

All text embeddings used for training and decoding were computed from the original Chinese stimulus sentences and Chinese semantic-category names. English translations are shown in figures and tables only for readability.

The resulting embedding is $\ell_2$-normalized to ensure consistent similarity measurements across samples. Crucially, the weights of $E_{\text{text}}$ were kept frozen throughout EEG encoder training. This defines a fixed semantic manifold that serves as a stable alignment target for the EEG encoder.  Given the limited vocabulary of the Chisco dataset, fine-tuning the language model could distort the semantic structure learned during large-scale pre-training, reduce inter-class separability, and increase overfitting risk \cite{kirkpatrick2017ewc,chen2020recall,dautume2019episodic}. A fixed text encoder also avoids the moving-target problem of joint optimization, ensuring that EEG representations are aligned to a consistent linguistic reference frame.

Given the fixed-encoder strategy, the structure of the semantic embedding space is fully determined by the choice of $E_{\text{text}}$. We therefore benchmarked several pretrained Chinese embedding models to identify an $E_{\text{text}}$ that provides a strong and stable target space for alignment while keeping the embedding dimensionality and model size practical.
Table~\ref{tab:text_encoder_models} summarizes the compared encoders together with their embedding dimensionality and parameter count.

\begin{table}[H]
\caption{Text Encoder Candidates}
\label{tab:text_encoder_models}
\centering
\setlength{\tabcolsep}{3pt}
\begin{tabular}{|p{165pt}|c|c|}
\hline
\textbf{Model} & \textbf{Dim.} & \textbf{Params} \\
\hline
BAAI/bge-small-zh-v1.5\cite{bge2023} & 512 & $\approx$24M \\
BAAI/bge-base-zh-v1.5\cite{bge2023} & 768 & $\approx$102M \\
BAAI/bge-large-zh-v1.5\cite{bge2023} & 1024 & $\approx$326M \\
intfloat/multilingual-e5-base\cite{e52024} & 768 & $\approx$278M \\
intfloat/multilingual-e5-large\cite{e52024} & 1024 & $\approx$560M \\
paraphrase-multilingual-mpnet-base-v2\cite{reimers2019sentencebert,song2020mpnet} & 768 & $\approx$278M \\
\hline
\end{tabular}
\end{table}

To evaluate the quality of the semantic space induced by each candidate model, we measured category separability for the 39 Chisco semantic classes. Because the EEG encoder is trained to map neural signals into this space, the separability of the text embeddings provides an empirical reference for how clearly the 39 intents are organized in the target language space.

We assessed separability using a suite of probing tasks, reporting Top-1 accuracy on the 39-way semantic category prediction task using: Linear Probe (Logistic Regression), Non-Linear Probes (MLP, XGBoost \cite{chen2016xgboost}) and Zero-Shot Baseline. 
These probing experiments quantify how well semantic categories are geometrically structured in the embedding manifold. In addition, text-encoder selection followed the cross-validation protocol, using only the training and validation partitions; the held-out test folds were not used.

\subsubsection{EEG Encoder}
\label{subsec:eeg_encoder}

We employ a trainable neural encoder $\mathcal{E}_{\text{eeg}}$ to project each imagined speech EEG trial $\mathbf{X} \in \mathbb{R}^{C \times T}$ (with $C=122$ channels and $T=1651$ time samples) into a semantic embedding $\mathbf{z} \in \mathbb{R}^{512}$, aligned with the frozen text encoder.

The EEG encoder follows a hybrid architecture with three stages: (i) a compact EEGNet-based convolutional backbone \cite{lawhern2018eegnet} for spatiotemporal feature extraction, producing a feature sequence $\mathbf{H}_{\mathrm{CNN}}$, (ii) a Transformer encoder \cite{vaswani2017transformer} for temporal modeling, implemented as a stack of 4 layers with 8 attention heads (key dimension 64), and (iii) a lightweight projection head.

In the final stage, the Transformer outputs are aggregated by attention pooling into a single vector, then mapped through a small MLP (256-unit hidden layer with GELU) to a 512-dimensional embedding. The output is $\ell_2$-normalized prior to computing the alignment losses to place EEG and text embeddings on a shared unit hypersphere, making similarity comparisons scale-invariant and stabilizing contrastive optimization.

\subsubsection{Training Objective}
\label{subsec:training}

The EEG encoder is optimized using a composite objective with two coupled aims:
(i) \textit{Semantic Alignment}, which encourages EEG embeddings to match the geometry of a frozen text embedding space, and
(ii) \textit{Discriminative Separability}, which preserves class-level separability across the 39 semantic categories.

The global loss function has the following form:
\begin{equation}
\mathcal{L}_{total} = \lambda_{align}\,\mathcal{L}_{SNCE} + \lambda_{cls}\,\mathcal{L}_{CE},
\label{eq:total_loss}
\end{equation}

where $\mathcal{L}_{SNCE}$ denotes a semantically reweighted InfoNCE-based alignment loss~\cite{infonce2018}, and $\mathcal{L}_{CE}$ denotes the weighted class-discrimination loss. The coefficients $\lambda_{align}$ and $\lambda_{cls}$ balance the relative contributions of alignment and class discrimination. In all experiments, we set $\lambda_{\text{align}} = 1$ and $\lambda_{\text{cls}} = 0.5$. These values were chosen in preliminary experiments on the basis of validation performance.

1) Semantic Alignment (Semantically Reweighted InfoNCE):
Let $\mathbf{t}_i \in \mathbb{R}^d$ be the frozen, $\ell_2$-normalized text embedding associated with sample $i$, and let
$\mathbf{z}_i \in \mathbb{R}^d$ be the corresponding $\ell_2$-normalized EEG embedding predicted by the encoder. The resulting soft-weighted contrastive loss is:
\begin{equation}
\mathcal{L}_{SNCE}
= -\frac{1}{N}\sum_{i=1}^{N} W_{y_i}\,
\log
\frac{\exp\!\big(s(\mathbf{z}_i,\mathbf{t}_i)/\tau\big)}
{\sum_{j=1}^{N} \omega_{i,j}\,\exp\!\big(s(\mathbf{z}_i,\mathbf{t}_j)/\tau\big)},
\label{eq:snce}
\end{equation}
where the temperature $\tau$ is initialized to $0.07$, optimized jointly with the EEG encoder, and constrained to the interval $[0.01,1.0]$.

For a mini-batch of size $N$, we construct a semantic reweighting matrix from text--text cosine similarities:
\begin{equation}
\omega_{i,j}
=
\begin{cases}
1, & i=j,\\
\max\!\left(\delta,1-s(\mathbf{t}_i,\mathbf{t}_j)\right),
& i\neq j,
\end{cases}
\label{eq:semantic_weights}
\end{equation}
where $s(\cdot,\cdot)$ denotes cosine similarity and $\delta = 0.05$ prevents near-zero weights. In contrast to standard InfoNCE, which treats all mismatched targets uniformly, the reweighting term $\omega_{i,j}$ modulates the contribution of each negative according to the semantic similarity between the correct text target $t_i$ and the competing target $t_j$. Consequently, negatives that are semantically close to $t_i$ are penalized less strongly, whereas semantically distant negatives receive larger weight. This yields a soft contrastive objective that better preserves the topology of the text space.

Because the dataset is class-imbalanced, we apply class-dependent sample weights inspired by the general motivation of class-balanced reweighting \cite{cui2019classbalanced}. Specifically, for each training sample $i$ with label $y_i$, we use an inverse square-root frequency heuristic, $W_{y_i} = (\mathrm{freq}_{y_i})^{-0.5}$, where $\mathrm{freq}_{y_i}$ denotes the number of training samples in class $y_i$ (computed within the fold-specific training pool). During optimization, each sample $i$ is weighted by $W_{y_i}$ in both $\mathcal{L}_{\mathrm{SNCE}}$ (Eq.~\ref{eq:snce}) and $\mathcal{L}_{\mathrm{CE}}$ (Eq.~\ref{eq:ce}).

2) Discriminative Regularization (Weighted Cross-Entropy):

The \textit{class discrimination loss}, $\mathcal{L}_{CE}$, is defined as
\begin{equation}
\scalebox{0.9}{$
\begin{aligned}
\mathcal{L}_{CE}
= -\frac{1}{N}\sum_{i=1}^{N} W_{y_i}
\sum_{c=1}^{C} y^{LS}_{i,c}\,
\log\left(
\frac{\exp\!\big(s(\mathbf{z}_i,\mathbf{p}_c)/\tau\big)}
{\sum_{c=1}^{C}\exp\!\big(s(\mathbf{z}_i,\mathbf{p}_c)/\tau\big)}
\right)
\end{aligned}
$},
\label{eq:ce}
\end{equation}
where $\mathbf{p}_c \in \mathbb{R}^{d}$ is the $\ell_2$-normalized prototype for class $c$, obtained by embedding the corresponding semantic category description shown in Fig.~\ref{fig:class_dist} using the same frozen text encoder used for $\mathbf{t}_i$. The same learned temperature $\tau$ used in the semantic alignment loss is applied to these prototype-based logits.

Rather than learning an unconstrained linear classification head, we tie the classification logits to frozen text-derived class prototypes $\mathbf{p}_c$. This step keeps the discriminative objective consistent with the semantic alignment loss, ensuring that supervision promotes class separability while remaining anchored to the geometry of the frozen language embedding space. We compute logits via scaled dot products and apply weighted cross-entropy with label smoothing ($\epsilon = 0.1$) \cite{szegedy2016inception}. In Eq.~\eqref{eq:ce}, the auxiliary term $y^{LS}_{i,c}$ is the smoothed target distribution, which is introduced to improve separability while remaining consistent with the frozen semantic prototypes.

\subsection{Task-Optimized Closed-Set Classifier}

For comparison, we implemented a conventional closed-set classifier optimized specifically for 39-way semantic-intent classification. The classifier received the same preprocessed imagined-speech EEG trials as the proposed framework and used the same within-subject cross-validation splits, validation partitions, held-out test trials, and random-seed protocol.

The classifier used a simplified EEGNet-like convolutional backbone consisting of temporal, depthwise spatial, and separable convolutional stages. The resulting feature sequence was combined with sinusoidal positional encoding and processed by a single Transformer encoder block with eight attention heads, a model dimension of 64, and a feed-forward dimension of 256. Global average pooling was then applied, followed by a 39-unit softmax classification layer. The model was optimized using weighted categorical cross-entropy with normalized inverse-square-root class-frequency weights. Unlike the proposed framework, this classifier does not use a text encoder or map EEG into a language embedding space. Because its architecture was optimized separately for closed-set classification, the comparison is system-level and does not isolate the effect of the training objective under an identical backbone.

\subsection{Evaluation Metrics and Analyses}
\label{subsec:evaluation_metrics}

Both decoding approaches are evaluated on the same held-out test folds using Top-$k$ accuracy, with $k \in \{1,5\}$, macro-F1, and balanced pairwise matching accuracy (2v2). Because the systems produce different output representations, 2v2 is computed from EEG--category-prototype similarities for the proposed framework and from class probabilities for the closed-set classifier, as described below. For the proposed framework, we additionally examine whether maximum EEG--text cosine similarity provides an intrinsic score related to prediction reliability, and probe semantic structure beyond the original category labels using evaluation-time semantic axes and the qualitative SGP.

\subsubsection{Balanced Pairwise Accuracy (2v2)}

In many practical BCI applications, the system often needs to discriminate between a small set of competing intents rather than the full vocabulary. We use the standard 2v2 accuracy \cite{mitchell2008nouns,sudre2012tracking,pereira2018universal,foster2021eeg2v2}, but estimate it using a class-balanced sampling strategy.

Specifically, we randomly sample two distinct semantic categories and then randomly select one held-out trial from each category, yielding trials $i$ and $j$ with labels $y_i\neq y_j$. Because the closed-set classifier predicts semantic categories rather than individual sentences, the proposed framework is also evaluated at the category level. Accordingly, its EEG embeddings are compared with the frozen category prototypes $\mathbf{p}_{y_i}$ and $\mathbf{p}_{y_j}$ defined in the language embedding space, rather than with the embeddings of the individual stimulus sentences. This ensures that both systems are evaluated on the same task: determining the correct category assignment for the two EEG trials.

For the proposed framework, a correct decision is made when the summed cosine similarity of the EEG embeddings to their corresponding category prototypes exceeds that of the swapped assignments:
\begin{align}
\scalebox{0.9}{$
\begin{gathered}
\text{Score}^{\mathrm{con}}_{2v2} =
\mathbb{I}\Big[
s(\mathbf{z}_i,\mathbf{p}_{y_i})
+s(\mathbf{z}_j,\mathbf{p}_{y_j})
>
s(\mathbf{z}_i,\mathbf{p}_{y_j})
+s(\mathbf{z}_j,\mathbf{p}_{y_i})
\Big],
\end{gathered}
$}
\end{align}
where  $s(\cdot,\cdot)$ denotes cosine similarity as before.

For the closed-set classifier, the corresponding comparison is performed using its predicted class probabilities. Let $P_i(c)$ denote the softmax probability assigned to category $c$ for trial $i$. The pairwise score is defined as
\begin{align}
\text{Score}^{\mathrm{closed}}_{2v2}
=
\mathbb{I}\Big[
P_i(y_i)+P_j(y_j)
>
P_i(y_j)+P_j(y_i)
\Big].
\end{align}

For both formulations, tied correct and swapped assignment scores were assigned half credit (0.5). We compute 2v2 separately for each test fold and report the mean $\pm$ standard deviation across the five folds. The theoretical chance level is 50\% for both formulations, making 2v2 an intuitive measure of two-way category-matching performance.

\subsubsection{Maximum Cosine Similarity Reliability Analysis}
\label{subsec:maxcosine_method}

For the proposed framework, category-level decoding is performed by comparing each EEG embedding with the frozen text-derived prototypes of the $C=39$ semantic categories. Because the predicted category is selected directly from these similarities, the similarity assigned to the selected category is immediately available without introducing a separate confidence-prediction head. We therefore examine whether this value provides an intrinsic signal related to decoding reliability.

For trial $i$, let $\mathbf{z}_i$ denote the $\ell_2$-normalized EEG embedding and let $\mathbf{p}_c$ denote the frozen, $\ell_2$-normalized prototype of category $c$. The predicted category $\hat{y}_i$ and the maximum cosine similarity score $m_i$ are defined as
\begin{equation}
\begin{aligned}
\hat{y}_i
&=
\underset{c \in \{1,\ldots,C\}}{\arg\max}\,
s(\mathbf{z}_i,\mathbf{p}_c), \\
m_i&=s(\mathbf{z}_i,\mathbf{p}_{\hat{y}_i}).
\end{aligned}
\label{eq:maxcosine}
\end{equation}
where $s(\cdot,\cdot)$ denotes cosine similarity. A larger $m_i$ indicates that the EEG embedding lies closer to its selected category prototype. However, $m_i$ is retained as a raw similarity score and is not interpreted as a calibrated probability.

We first examine whether validation-derived MaxCosine thresholds transfer consistently to held-out data. Within each fold, we determined thresholds corresponding to target validation coverage levels of $100\%$, $80\%$, $60\%$, $40\%$, and $20\%$, using only the validation-score distribution. Each threshold is then applied unchanged to the corresponding held-out test fold, and the resulting test coverage is compared with the target coverage used to derive the threshold. Close agreement between target and test coverage indicates that the relevant quantiles of the MaxCosine distribution are reasonably stable across validation and test folds, although it does not imply that the full score distributions are identical.

Using these validation-derived thresholds, we then examine selective decoding performance by reporting the achieved test coverage and the Top-1 and Top-5 accuracies among the retained trials. For an additional visualization of the relationship between MaxCosine and decoding accuracy, each test MaxCosine score is converted to its percentile rank within the corresponding fold-specific validation distribution and then assigned to one of five non-overlapping quintiles.

Finally, we quantify the ability of $m_i$ to discriminate correct from incorrect Top-1 predictions using the area under the receiver operating characteristic curve (AUROC), with correctness defined as $\mathbb{I}[\hat{y}_i=y_i]$. AUROC is computed separately for each held-out test fold and summarized using the mean $\pm$ standard deviation.

Test labels are used only to evaluate correctness after the thresholds and quintile boundaries have been established. Collectively, these analyses assess whether MaxCosine provides a useful ranking signal related to prediction reliability. They do not establish probabilistic calibration or operational online performance.

\subsubsection{Generalization Across Semantic Axes}
\label{subsec:semantic_axes_method}

As an exploratory post hoc analysis, we examine whether the learned EEG embeddings preserve broader semantic structure beyond the original 39-category taxonomy. To do so, we define five bipolar semantic axes directly in the frozen language embedding space. These axes are introduced only at evaluation time and are therefore not used as labels or supervision during EEG-encoder training. No additional classifiers are trained for this analysis.

Each pole of an axis is represented by three manually defined Chinese anchor phrases describing the corresponding semantic concept. The phrases are embedded using the same frozen text encoder employed throughout the proposed framework. For each pole, the three resulting embeddings are averaged and $\ell_2$-normalized to obtain a single anchor vector. Table~\ref{tab:semantic_axes} summarizes the five semantic axes using their anchor concepts for readability, while the original Chinese anchor phrases are used for all computations.

\begin{table}[H]
\caption{Semantic axes used in the evaluation-time generalization analysis.}
\label{tab:semantic_axes}
\centering
\setlength{\tabcolsep}{3pt}
\begin{tabular}{|p{20pt}|p{105pt}|p{100pt}|}
\hline
\textbf{Axis} & \textbf{Left pole} & \textbf{Right pole} \\
\hline
A1 & Practical/Transactional & Social/Emotional \\
A2 & Physical/Embodied & Informational/Cognitive \\
A3 & Goal-directed/Obligation & Leisure/Enjoyment \\
A4 & Mobility/Outside-world & Home/Personal-life \\
A5 & Individual/Self-related & Interpersonal/Social \\
\hline
\end{tabular}
\end{table}

For each semantic axis $k$, let $\mathbf{a}^{L}_{k}$ and $\mathbf{a}^{R}_{k}$ denote the normalized anchor vectors corresponding to its left and right poles. For a representation $\mathbf{v}$, its position on the axis is determined by comparing its cosine similarity to the two anchor vectors:
\begin{equation}
q_k(\mathbf{v}) =
\begin{cases}
L, & s(\mathbf{v},\mathbf{a}^{L}_{k}) >
     s(\mathbf{v},\mathbf{a}^{R}_{k}),\\
R, & \text{otherwise},
\end{cases}
\label{eq:semantic_axis_assignment}
\end{equation}
where $s(\cdot,\cdot)$ denotes cosine similarity.

For every held-out test trial, this assignment is performed independently for the EEG embedding $\mathbf{z}_i$ and for the frozen text embedding $\mathbf{t}_i$ of its corresponding imagined sentence. The sentence-derived assignment $q_k(\mathbf{t}_i)$ serves as the semantic reference and is compared with the EEG-derived assignment $q_k(\mathbf{z}_i)$. In this way, the analysis tests whether the EEG representation places a held-out trial on the same side of a newly introduced semantic dimension as its corresponding sentence representation.

Agreement is quantified using balanced accuracy, which gives equal weight to the two sides of each semantic axis and has a chance level of 50\%. For each subject and semantic axis, balanced accuracy is computed over all held-out test trials collected across the five cross-validation folds. The resulting subject-level scores are then summarized across the five participants as the mean $\pm$ standard deviation.

\subsubsection{Semantic Gallery Probe (SGP)}
\label{subsec:semantic_spotlight_method}

Complementing the quantitative semantic-axis analysis, we use SGP to qualitatively examine whether the learned EEG embeddings extend beyond the \emph{exact} training sentences. Because the proposed framework maps EEG and text into a shared frozen language space, previously unseen candidate sentences can be compared directly with a held-out EEG embedding without retraining the EEG encoder or modifying its output layer. This capability is not directly available to the closed-set classifier, whose outputs are restricted to the fixed set of 39 semantic categories.

We construct a frozen semantic gallery of manually written, held-out sentences that express the same target concepts using alternative phrasings (e.g., additional paraphrases within ``Food''). All gallery sentences are embedded with the frozen text encoder and projected to 2D via PCA for visualization. Given a single EEG test trial, we visualize how it relates to this unseen gallery by coloring each gallery point according to its cosine similarity to the EEG embedding. For qualitative illustration, we report a representative held-out trial with a high MaxCosine score. We inspect whether the EEG embedding assigns relatively higher similarity to candidate sentences in the intended semantic region, including sentences not observed during EEG-encoder training, while assigning lower similarity to unrelated regions. SGP is therefore used as a qualitative demonstration of candidate-set flexibility rather than as a quantitative performance metric. Visualizations are reported in Section~\ref{sec:results}.

\section{Experimental Setup}
\label{sec:experiments}

\subsection{Protocol and Reporting}

We use stratified, sentence-grouped five-fold cross-validation with a fixed random seed, ensuring that all trials associated with the same sentence remain in the same fold. In each fold, one fold is held out for testing, while $15\%$ of the sentence groups in the remaining training pool are reserved for validation. The same splits are used throughout the study. Held-out test data are used only for final model evaluation and are excluded from all model-selection procedures.

\subsection{Training Configuration}

Both decoding approaches were implemented in TensorFlow and optimized with Adam~\cite{kingma2014adam} for a maximum of 100 epochs. Because the proposed framework and the closed-set classifier were optimized for different objectives, they used task-specific training and model-selection configurations.

The proposed contrastive framework was trained with a batch size of 64 and an initial learning rate of $3\times10^{-4}$. The learning rate was multiplied by $0.7$ after five epochs without improvement in validation loss, down to a minimum of $1\times10^{-6}$. Early stopping was applied with a patience of 16 epochs. During training, we applied additive Gaussian noise ($\sigma=0.05$), channel dropout that randomly removed $10\%$ of the channels with probability $0.1$, and temporal shifts implemented as circular rolls by $\delta\in[-50,50]$ samples.

The closed-set classifier was trained with a batch size of 16 and an initial learning rate of $5\times10^{-4}$. Its learning rate was multiplied by $0.7$ after four epochs without improvement in validation loss, down to a minimum of $1\times10^{-6}$, and early stopping was applied with a patience of nine epochs. No additional EEG augmentation was applied during closed-set classifier training.

For both systems, learning-rate scheduling and early stopping were performed independently within each cross-validation fold. The validation Top-5 criterion was used for the proposed contrastive framework because its decoding objective emphasizes ranked retrieval, whereas validation Top-1 was used for the closed-set classifier because it was optimized for exact category classification. These model-selection criteria were fixed before evaluation on the held-out test folds. Each selected checkpoint was subsequently evaluated on the corresponding test fold, and both systems were evaluated using the same held-out test samples.

\subsection{Chance Baselines and Sampling Procedures}

Top-k accuracy and macro-F1 are compared against a stratified random baseline that reflects the class distribution of each fold-specific training set. Chance performance is estimated using $10{,}000$ Monte Carlo samples per fold.

Balanced pairwise accuracy (2v2) is estimated separately for each held-out test fold using $20{,}000$ class-balanced pairs. The same sampling procedure is applied to both systems.

\section{Results}
\label{sec:results}

\subsection{Text Encoder Selection}
The choice of text-embedding model directly shapes the semantic geometry for the EEG representations, and it provides a practical reference for how separable the intent categories are in the target space. We therefore first report the performances  of candidate language embedding models on the 39 Chisco semantic categories, using probes trained \emph{only} on text embeddings. This analysis serves two purposes: (i) it identifies a target semantic space that is sufficiently structured for the downstream alignment task, and (ii) it provides an empirical reference for the category separability of the frozen text space.

\begin{figure}[H]
\centerline{\includegraphics[width=\columnwidth]{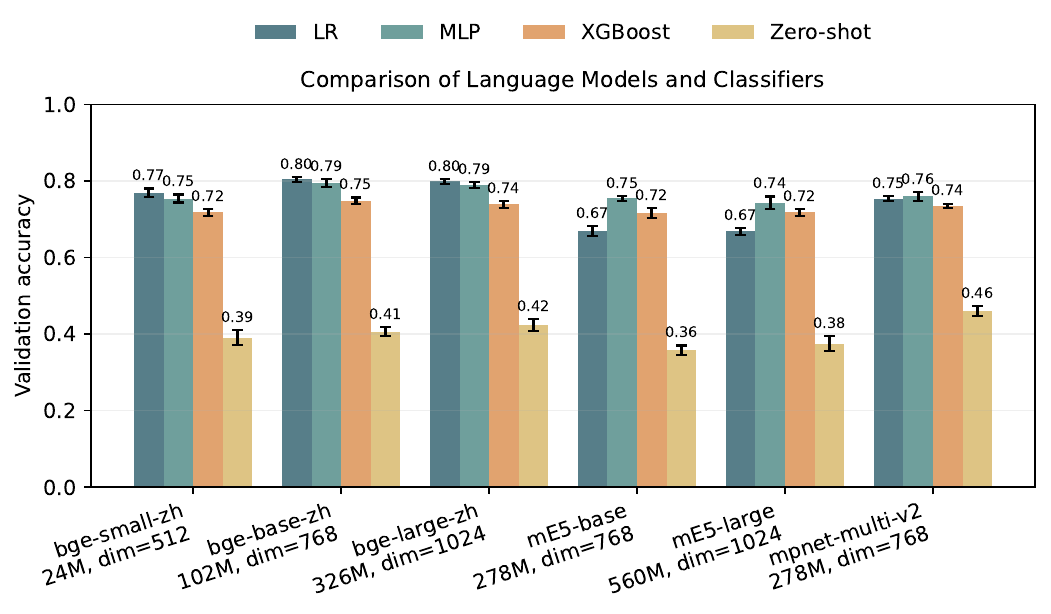}}
\caption[Text encoder probe accuracy across models]{Empirical comparison of different Chinese language embedding models on the Chisco dataset. Bars represent classification Top-1 accuracy of probes (LR, MLP, XGBoost) trained on the text embeddings, and a zero-shot baseline computed without supervision. Error bars indicate ± std across folds.}
\label{fig:text_encoders}
\end{figure}

Figure~\ref{fig:text_encoders} compares the candidate encoders listed in Table~\ref{tab:text_encoder_models} using mean validation Top-1 accuracy across the five folds. Performance did not increase uniformly with model size. The bge-base and bge-large encoders achieved the highest linear-probe accuracy of approximately 0.80, with MLP accuracy of approximately 0.79. The compact bge-small encoder achieved accuracies of approximately 0.77, 0.75, and 0.72 with the linear, MLP, and XGBoost probes, respectively, while using only 24M parameters and 512-dimensional embeddings. Given this trade-off, we selected BAAI/bge-small-zh-v1.5 as the frozen text encoder for the subsequent EEG--text alignment experiments, balancing semantic separability with computational efficiency and memory requirements.

\subsection{Classification and Discriminability Metrics}

\begin{figure}[H]
\centering
\centerline{\includegraphics[width=\columnwidth]{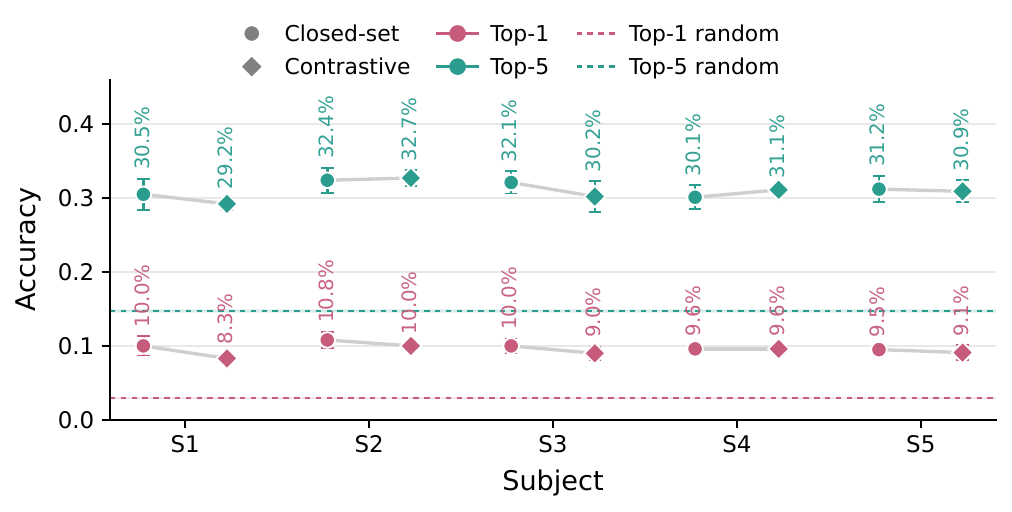}}
\centerline{\footnotesize (a) Subject-wise Top-$k$ decoding accuracy.}
\vspace{0.5cm}
\centerline{\includegraphics[width=\columnwidth]{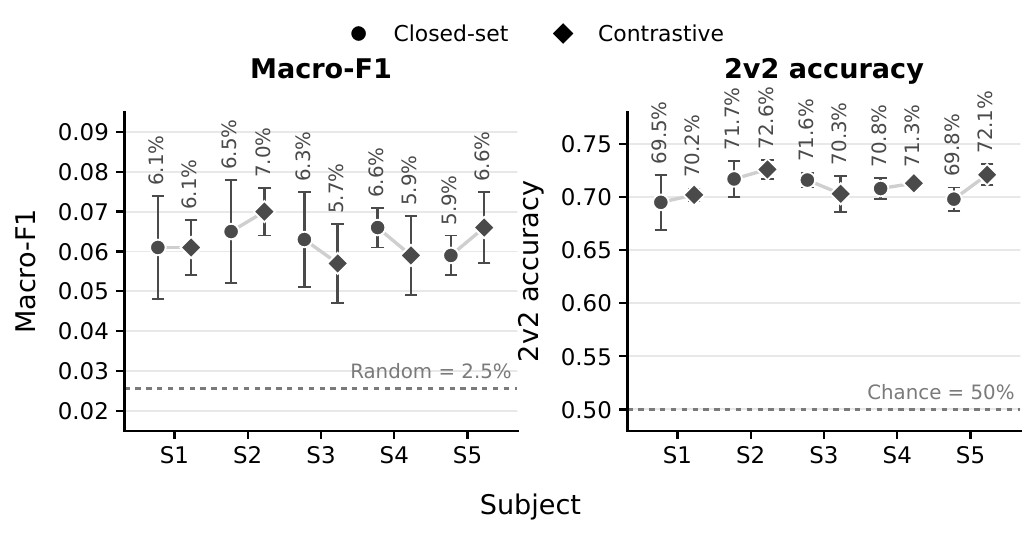}}
\centerline{\footnotesize (b) Subject-wise Macro-F1 and balanced 2v2 accuracy.}
\caption[Top-$k$, Macro-F1, and 2v2 across subjects]{Classification performance of the task-optimized closed-set classifier and the proposed contrastive EEG--language framework across Subjects 01--05. Results are reported as the mean $\pm$ std across the five held-out cross-validation folds.}
\label{fig:metrics_all}
\end{figure}

Fig.~\ref{fig:metrics_all}(a) compares the Top-$k$ decoding performance of the two systems across the five subjects. Both systems remained above the prior-aware random baseline at every reported value of $k$. Across subjects and systems, mean Top-1 accuracy was approximately 0.08--0.11 (chance level 0.03), increasing to approximately 0.29--0.33 (chance level 0.15), at Top-5. Performance was closely matched between the two systems across subjects, showing that mapping EEG into the frozen language space preserved conventional Top-$k$ decoding performance relative to the task-optimized closed-set classifier.

Fig.~\ref{fig:metrics_all}(b) presents Macro-F1 and balanced category-level 2v2 accuracy. Macro-F1 remained above the prior-aware random baseline of approximately 0.025 for both systems and all subjects, with mean values ranging from approximately 0.057 to 0.070. The two systems showed closely matched Macro-F1 performance overall, with small subject-specific differences: the contrastive framework was higher for Subjects 02 and 05, the closed-set classifier was higher for Subjects 03 and 04, and the two were similar for Subject 01.

Both systems achieved balanced 2v2 accuracies clearly above the theoretical chance level of 0.50, with subject-level means of approximately 0.69--0.73. The contrastive framework was numerically higher for four of the five subjects, while the closed-set classifier was higher for Subject 03. Overall, the two systems showed comparable pairwise discriminability, providing further evidence that the language-aligned representation preserves conventional category-level decoding performance.

\subsection{Maximum Cosine Similarity Reliability Analysis}
\label{subsec:maxcosine_results}

We next examine whether the maximum EEG--text cosine similarity assigned to the predicted category provides an intrinsic signal related to decoding reliability. Subject 01, the first indexed participant, is presented as an illustrative example, followed later by a summary across all subjects. Because this score is computed directly from the similarities already used for decoding, it is available for every trial at inference time without requiring a separate confidence-prediction head or access to the true label. We therefore examine whether MaxCosine provides an intrinsic ranking signal for prediction reliability that is available directly at inference time. Throughout this analysis, MaxCosine is treated as a raw similarity-based reliability score rather than as a calibrated probability.

\subsubsection{Validation-to-Test Coverage Transfer}

\begin{figure}[H]
\centering
\centerline{\includegraphics[width=\columnwidth]{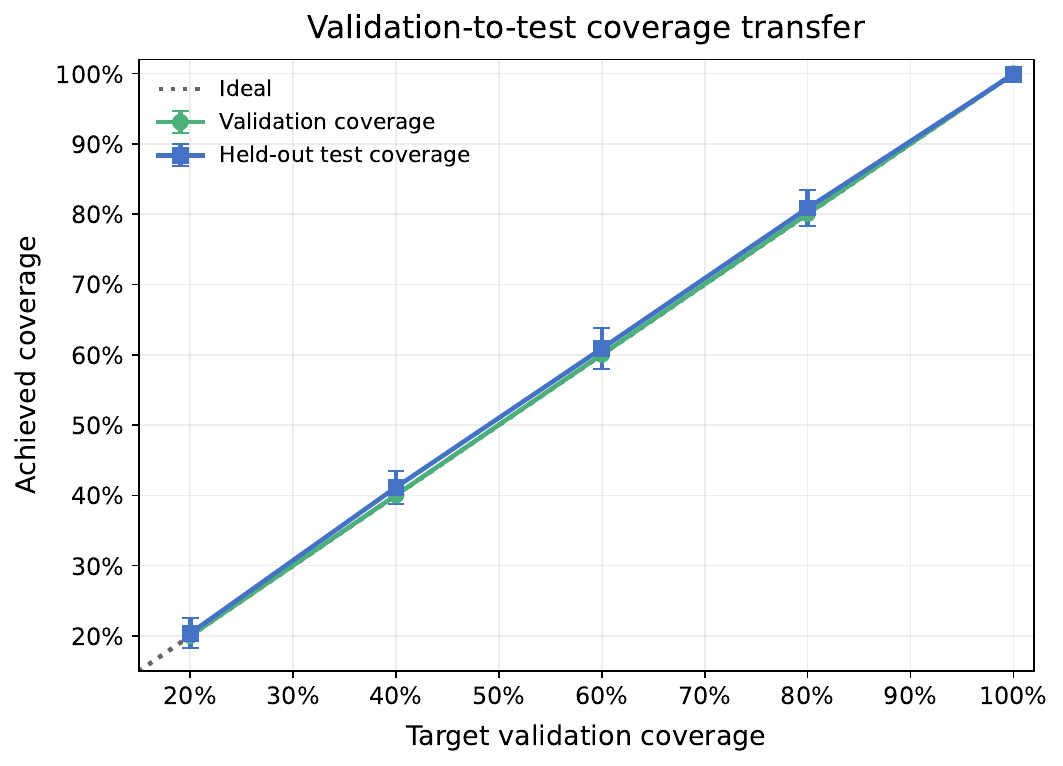}}
\caption[Validation-to-test coverage transfer]{Validation-to-test transfer of MaxCosine thresholds for Subject 01. Fold-specific thresholds corresponding to the target coverage levels were determined exclusively from the validation-score distribution and subsequently applied unchanged to the corresponding held-out test fold. Points and error bars represent the mean $\pm$ std across the five folds.}
\label{fig:maxcosine_coverage_transfer}
\end{figure}

Fig.~\ref{fig:maxcosine_coverage_transfer} shows the achieved validation and held-out test coverage at each target validation coverage level. As expected from the threshold-selection procedure, achieved validation coverage closely followed the specified targets. More importantly, the corresponding held-out test coverage remained close to the validation coverage across the examined operating points. This agreement indicates that the relevant MaxCosine quantiles were stable between the validation and test partitions for Subject 01, allowing validation-derived operating thresholds to transfer without using test labels. The analysis therefore supports threshold transfer at the examined operating points, without requiring equality of the complete validation and test score distributions.

\subsubsection{Selective Accuracy Across Coverage Levels}

\begin{figure}[H]
\centering
\centerline{\includegraphics[width=\columnwidth]{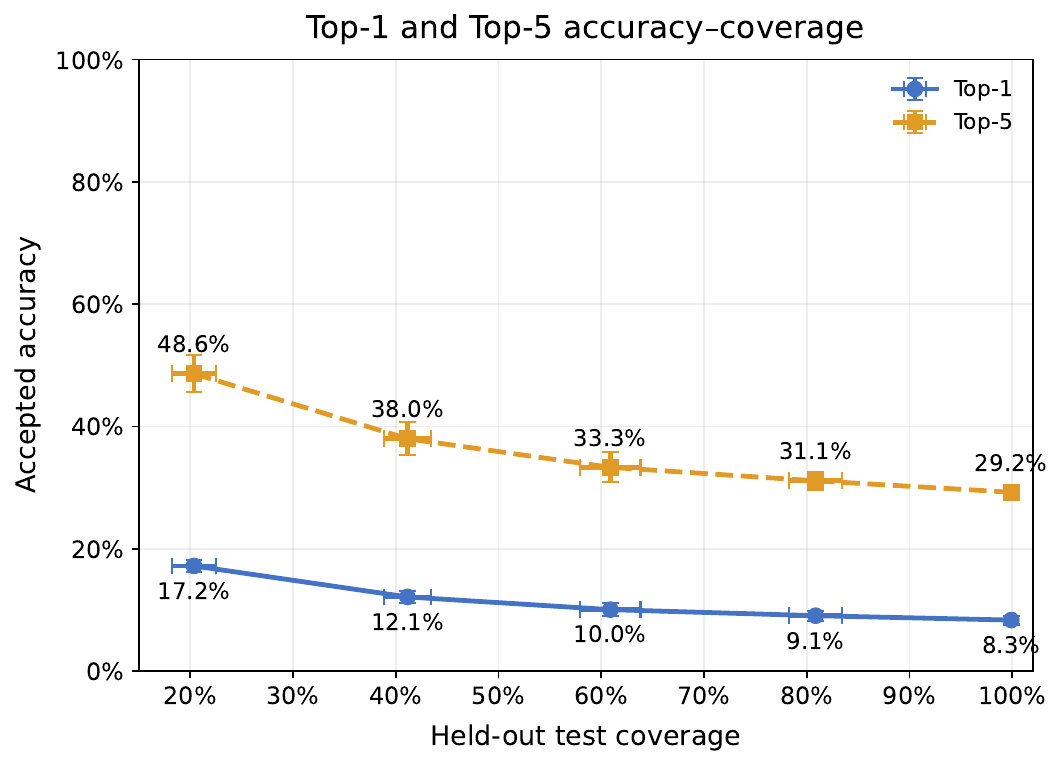}}
\caption[Selective accuracy across coverage levels]{Top-1 and Top-5 accuracy among the held-out test trials retained by validation-derived MaxCosine thresholds for Subject 01. The horizontal axis reports the achieved held-out test coverage. Points and error bars represent the mean $\pm$ standard deviation across the five folds.}
\label{fig:maxcosine_accuracy_coverage}
\end{figure}

Fig.~\ref{fig:maxcosine_accuracy_coverage} shows the relationship between retained test coverage and decoding accuracy. At approximately 100\% coverage, mean Top-1 and Top-5 accuracy were approximately 8.3\% and 29\%, respectively. As coverage decreased, both measures generally increased. At approximately 20\% coverage, Top-1 accuracy reached 17.2\%, while Top-5 accuracy approached 49\%. At the validation coverage targets of 80\%, 60\%, and 40\%, accuracy also increased as fewer test trials were retained, indicating that higher validation-derived MaxCosine thresholds selected subsets with a greater proportion of correct predictions. These results demonstrate that MaxCosine provides a useful offline ranking signal for selective decoding. Calibration and online abstention are separate operational questions and are not evaluated in the present experiment.

\subsubsection{Accuracy Across MaxCosine Quintiles}

\begin{figure}[H]
\centering
\centerline{\includegraphics[width=\columnwidth]{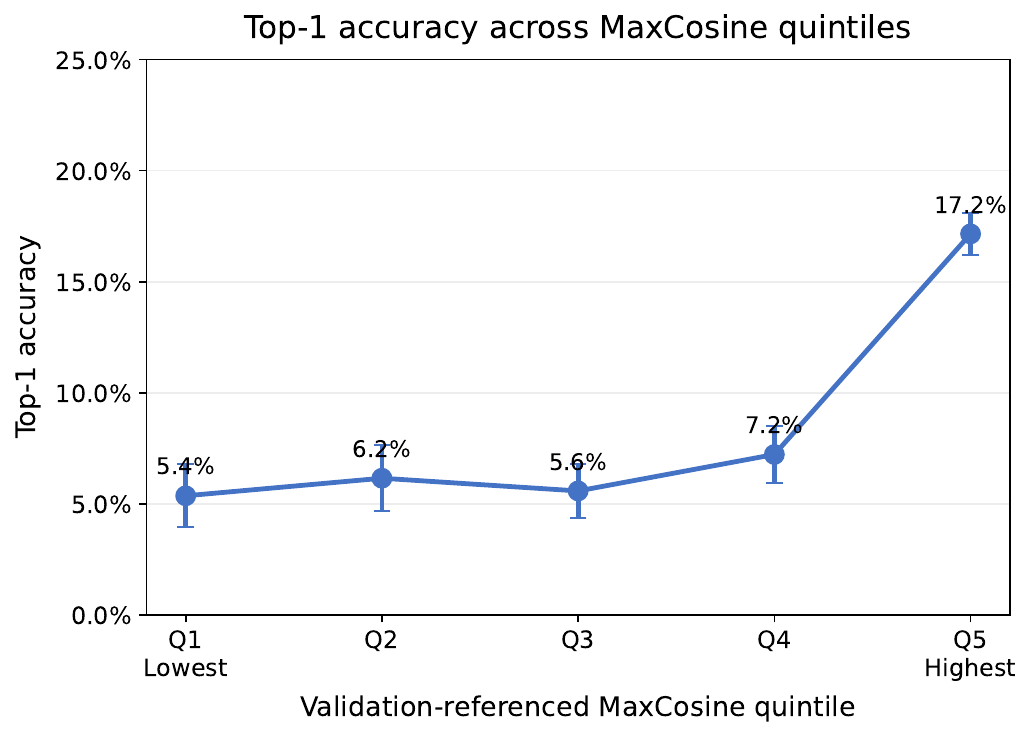}}
\caption[Top-1 accuracy across MaxCosine quintiles]{Top-1 accuracy across five non-overlapping, validation-referenced MaxCosine quintiles for Subject 01. Each held-out test score is expressed relative to its fold-specific validation-score distribution and assigned to one of five quintiles, from the lowest (Q1) to the highest (Q5). Points and error bars represent the mean $\pm$ standard deviation across the five folds.}
\label{fig:maxcosine_quintiles}
\end{figure}

Fig.~\ref{fig:maxcosine_quintiles} examines Top-1 accuracy across five non-overlapping, validation-referenced MaxCosine quintiles. The highest MaxCosine quintile exhibited a clear separation, reaching a mean Top-1 accuracy of 17.2\%, compared with 5.4\%--7.2\% across the lower four quintiles. The lower quintiles showed comparatively similar accuracies without a strict ordering. Thus, the most pronounced reliability information was concentrated in the highest-scoring MaxCosine quintile, consistent with the selective accuracy--coverage analysis. Unlike the cumulative accuracy--coverage analysis, each trial contributes to only one quintile in this analysis.

\subsubsection{Illustrative Predictions Across MaxCosine Percentiles}

\begin{table}[!b]
\caption[Illustrative predictions across MaxCosine percentiles]{Illustrative held-out Top-5 decoding examples across validation-referenced MaxCosine percentiles for Subject 01. Percentiles express each test score relative to its fold-specific validation-score distribution. A higher percentile therefore indicates a higher relative MaxCosine score, not a calibrated probability of correctness.}
\label{tab:maxcosine_examples}
\centering
\scriptsize
\setlength{\tabcolsep}{3pt}
\renewcommand{\arraystretch}{1.2}
\begin{tabular}{|p{55pt}|p{158pt}|}
\hline
\textbf{Validation-referenced percentile} &
\textbf{Held-out test example} \\
\hline

\textbf{P100 target} &
``I am interested in the field of marketing."\newline
\textbf{True category:} Job Hunting and Career Development\newline
\textbf{Top-5 predictions:} Job Hunting and Career Development; Work; Learning; Personal Information; Education\\
\hline

\textbf{P90 target} &
``Deposits, Withdrawals and Checking Accounts."\newline
\textbf{True category:} Finance\newline
\textbf{Top-5 predictions:} Finance; Prices and Costs; Personal Information; Inquiries and Personal Matters; Providing and Requesting Assistance\\
\hline

\textbf{P80 target} &
``Books borrowed from the library can be renewed."\newline
\textbf{True category:} Internet and Information Technology\newline
\textbf{Top-5 predictions:} Education; Learning; Products and Quality Assurance; Food and Dining; Welcoming and Thanking\\
\hline
\end{tabular}
\end{table}

To complement the aggregate reliability analyses, Table~\ref{tab:maxcosine_examples} presents three illustrative held-out trials from Subject 01 selected near three predefined validation-referenced MaxCosine percentile targets, 100\%, 90\%, and 80\%. For each target, a held-out test trial with a validation-referenced percentile close to the specified value was selected for qualitative inspection. For each trial, the table reports the input sentence, true semantic category, and five highest-ranked category predictions.

The three examples illustrate how the ranked output can be inspected at different high MaxCosine levels. For the P100-target example, ``I am interested in the field of marketing,'' the correct category, Job Hunting and Career Development, was ranked first. For the P90-target example, ``Deposits, Withdrawals and Checking Accounts,'' the correct Finance category was also ranked first. For the P80-target example, ``Books borrowed from the library can be renewed,'' the true Internet and Information Technology category was absent from the five highest-ranked predictions. These examples are intended only to make the model outputs and reliability ranking concrete; conclusions about the overall relationship between MaxCosine and correctness are based on the coverage, quintile, AUROC, and across-subject analyses rather than on these individual trials.

\subsubsection{Discrimination of Prediction Correctness}

\begin{figure}[H]
\centering
\centerline{\includegraphics[width=\columnwidth]{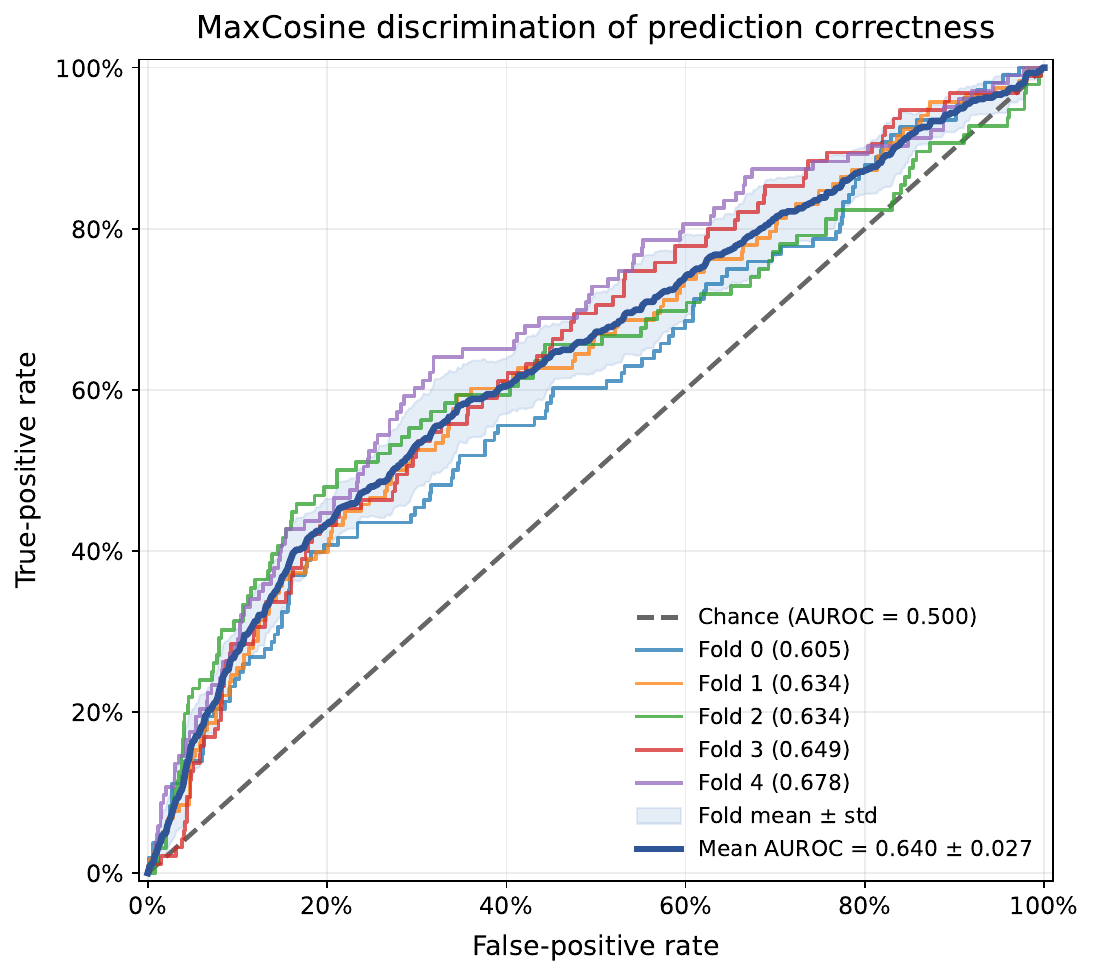}}
\caption[ROC analysis of MaxCosine and prediction correctness]{Fold-specific ROC curves for discrimination of correct and incorrect Top-1 predictions using raw MaxCosine for Subject 01.}
\label{fig:maxcosine_roc}
\end{figure}

Fig.~\ref{fig:maxcosine_roc} evaluates the ability of raw MaxCosine to discriminate correct from incorrect Top-1 predictions. Fold-specific AUROC values ranged from 0.605 to 0.678, and the mean AUROC across the five folds was $0.640 \pm 0.027$. All five fold-specific AUROC values were above the chance reference of 0.50, indicating consistent above-chance discrimination between correct and incorrect Top-1 predictions for Subject 01. The mean AUROC of $0.640 \pm 0.027$ supports MaxCosine as a useful reliability-ranking signal.

\subsubsection{Across-Subject Summary}

The preceding analyses provide a detailed illustration of MaxCosine behavior for Subject 01. To determine whether the observed pattern is specific to this example or is also present across participants, Table~\ref{tab:maxcosine_subjects} reports the principal subject-level results. The table summarizes correctness discrimination using AUROC, validation-to-test coverage transfer at the 20\% target operating point, and the corresponding changes in Top-1 and Top-5 accuracy between approximately full coverage and selective retention.

\begin{table}[H]
\centering
\caption[Across-subject MaxCosine reliability summary]{Across-subject summary of the MaxCosine reliability analysis. AUROC is reported as the mean $\pm$ standard deviation across the five folds. ``Coverage at 20\%'' denotes the mean achieved held-out test coverage after applying the fold-specific threshold targeting 20\% validation coverage. The Top-1 and Top-5 columns report mean test accuracy at approximately full coverage and at the 20\% validation-target operating point, respectively. Coverage and accuracy values are expressed as percentages.}
\label{tab:maxcosine_subjects}
\resizebox{\columnwidth}{!}{%
\begin{tabular}{lccccc}
\hline
\textbf{Subject}
&
\shortstack{\textbf{AUROC}\\\textbf{mean $\pm$ std}}
&
\shortstack{\textbf{Coverage}\\\textbf{at 20\%}}
&
\shortstack{\textbf{Top-1 accuracy}\\\textbf{100\% $\rightarrow$ 20\%}}
&
\shortstack{\textbf{Top-5 accuracy}\\\textbf{100\% $\rightarrow$ 20\%}}

\\
\hline
S1 & 0.64 $\pm$ 0.027 & 20.4 & 8.3 $\rightarrow$ 17.2 & 29.2 $\rightarrow$ 48.6 \\
S2 & 0.61 $\pm$ 0.027 & 19.3 & 10 $\rightarrow$ 19.2 & 32.7 $\rightarrow$ 50.5\\
S3 & 0.60  $\pm$ 0.020& 19.7 & 9 $\rightarrow$ 15.6 & 30.1 $\rightarrow$ 42.9 \\
S4 & 0.62  $\pm$ 0.027& 20 & 9.6 $\rightarrow$ 18.3 & 31.1 $\rightarrow$ 46.2 \\
S5 & 0.61  $\pm$ 0.058& 19.8 & 9.1 $\rightarrow$ 16.9 & 30.9 $\rightarrow$ 45.9 \\
\hline
\end{tabular}%
}
\end{table}

Across all five subjects, mean AUROC remained above the 0.50 chance reference, showing that the relationship between MaxCosine and prediction correctness was consistent across participants. Validation-derived thresholds targeting 20\% coverage transferred closely to the held-out test folds, with achieved test coverage remaining near the intended operating point. More importantly, retaining the highest-MaxCosine trials increased both Top-1 and Top-5 accuracy for every subject. Top-1 accuracy increased from 8.3--10.0\% at full coverage to 15.6--19.2\% at the selective operating point, while Top-5 accuracy increased from 29.2--32.7\% to 42.9--50.5\%. Thus, the detailed pattern observed for Subject 01 generalized across all five participants: higher MaxCosine scores consistently identified smaller subsets of trials with greater decoding accuracy. The AUROC values indicate informative, although incomplete, reliability ranking and are interpreted independently from probabilistic calibration.

\subsection{Generalization Across Semantic Axes}
\label{subsec:semantic_axes_results}

\begin{figure}[H]
\centerline{\includegraphics[width=\columnwidth]{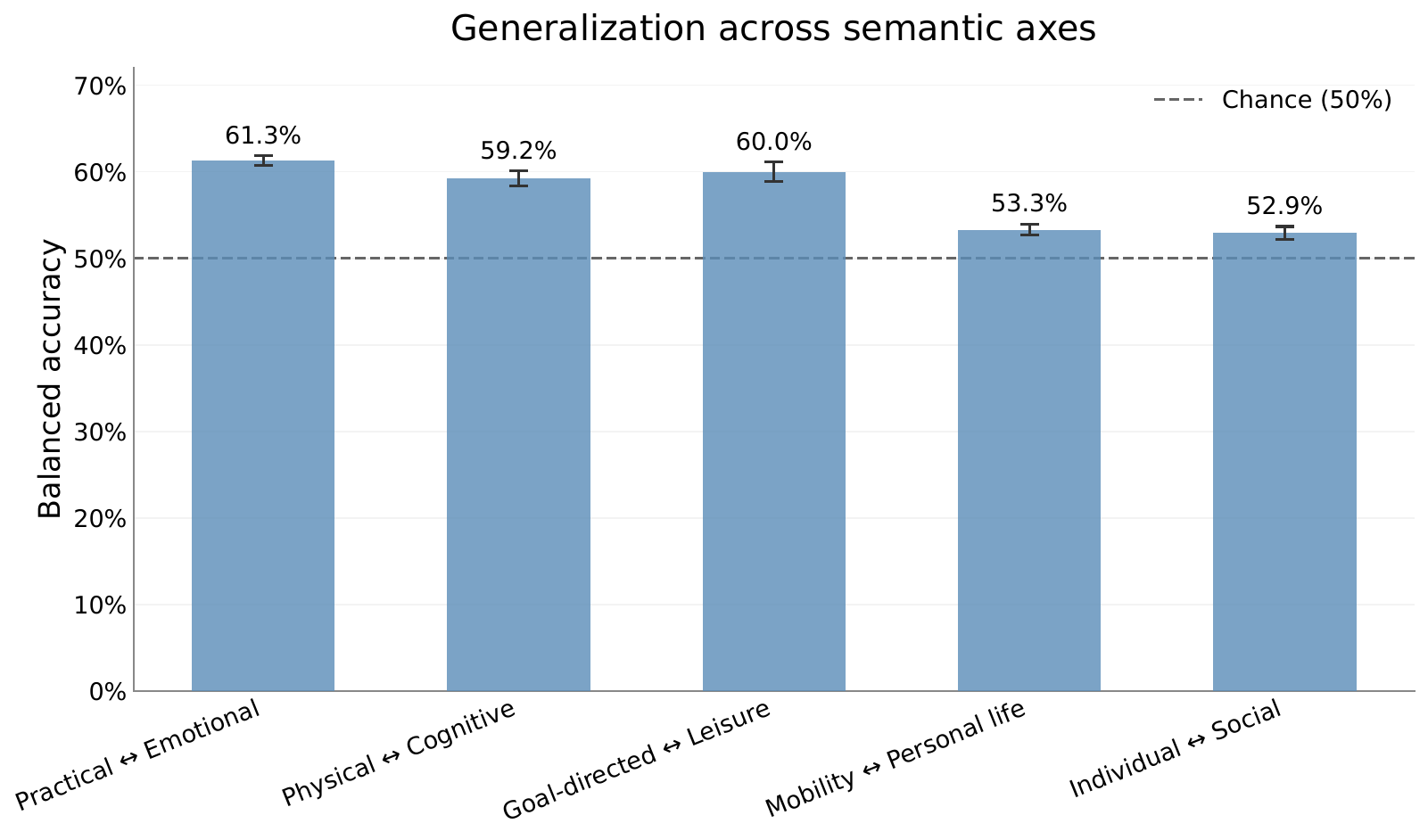}}
\caption[Generalization across semantic axes]{Balanced accuracy measures agreement between the semantic-axis assignment derived from the held-out EEG embedding and that derived from the corresponding sentence embedding. Bars show the mean across the five subjects and error bars indicate the standard deviation across subjects. The dashed line denotes the 50\% chance level.}
\label{fig:semantic_axes}
\end{figure}

Figure~\ref{fig:semantic_axes} summarizes generalization across the five evaluation-time semantic axes. Balanced accuracy was computed separately for each subject using all held-out test trials across the five cross-validation folds and was then summarized across the five subjects as the mean $\pm$ std.

Mean balanced accuracy was above the 50\% chance level for all five axes. The strongest agreement was observed for Practical--Emotional (61.3\%), followed by Goal-directed--Leisure (60.0\%) and Physical--Cognitive (59.2\%).

These results indicate that the learned EEG embeddings preserve broader semantic organization represented in the frozen language space, including along dimensions introduced only at evaluation time and never used as training labels. The strength of this correspondence varied across dimensions, with the clearest agreement observed for Practical--Emotional, Goal-directed--Leisure, and Physical--Cognitive.

\subsection{Semantic Gallery Probe (SGP)}
\label{subsec:neural_spotlight_results}

\begin{figure}[H]
\centerline{\includegraphics[width=\columnwidth]{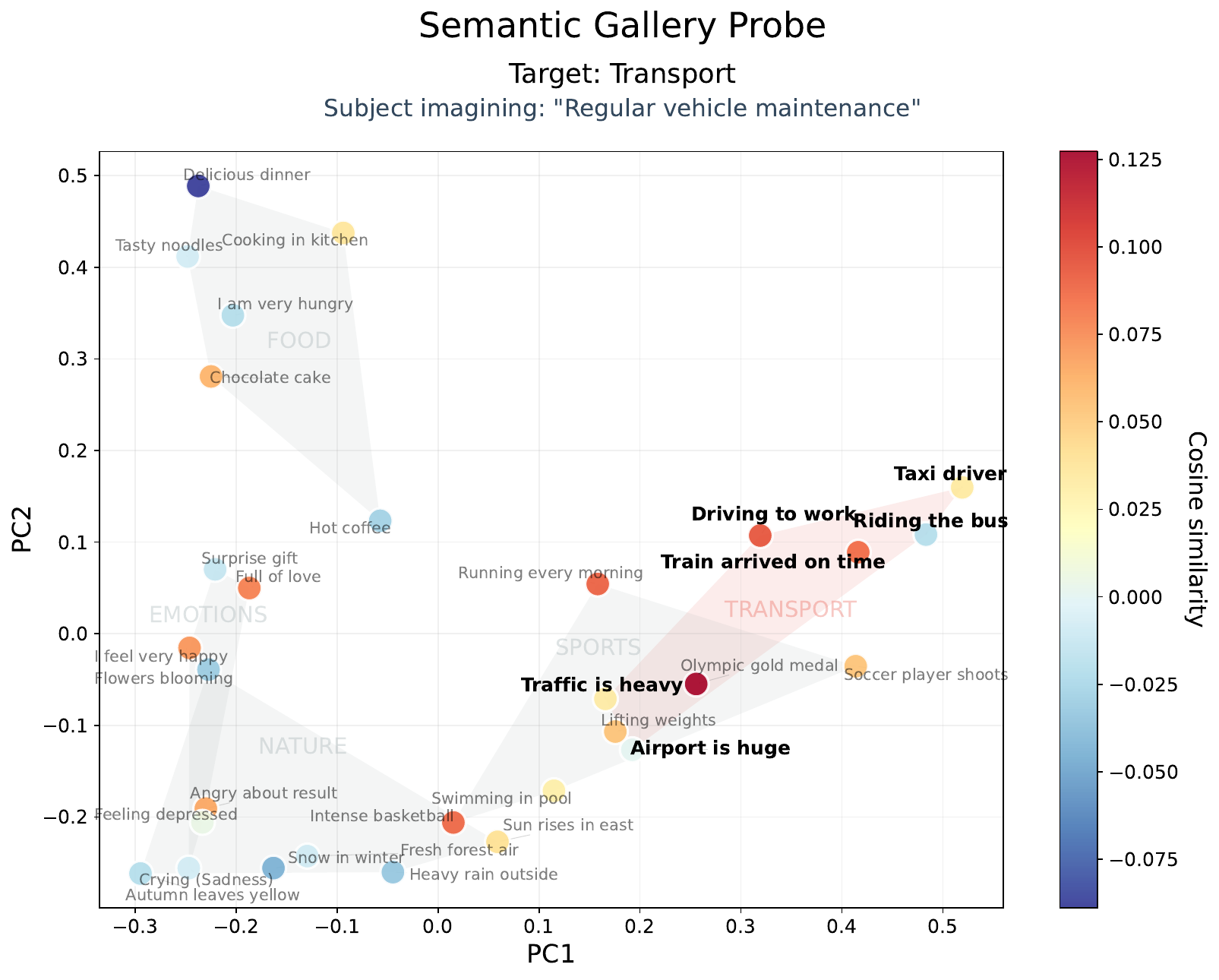}}
\caption[Semantic Gallery Probe (SGP) visualization for Subject 1]{SGP example for Subject~1. A frozen gallery of held-out sentence embeddings is projected to 2D via PCA for visualization. Each point corresponds to a gallery sentence and is colored by its cosine similarity to the EEG trial.}
\label{fig:neural_spotlight_s1}
\end{figure}

Complementing the quantitative semantic-axis analysis, we use \emph{SGP} to qualitatively examine how a single-trial EEG embedding aligns within the frozen language space. SGP compares a held-out EEG test trial to a fixed semantic gallery of manually written sentences that were not used during training, providing a qualitative visualization of where the EEG embedding places similarity mass within the frozen text space.

Figure~\ref{fig:neural_spotlight_s1} shows a representative held-out test trial where similarity concentrates on a localized region of the gallery. In this example, sentences associated with the intended Transport-related context receive elevated similarity, whereas semantically distant regions, such as Nature, remain comparatively low.

We emphasize that SGP is a qualitative visualization rather than a standalone performance metric or evidence of open-vocabulary generalization. Its purpose is to illustrate where a trial places similarity mass within an unseen semantic gallery. Similar qualitative patterns were observed across subjects.

\section{Discussion}
\label{sec:discussion}

The present results show that contrastive EEG--language alignment can preserve conventional imagined-speech decoding performance while changing the form of the learned representation. Across the five subjects, the proposed framework achieved Top-$k$, macro-F1, and pairwise discriminability comparable to the task-optimized closed-set classifier, while mapping single-trial EEG into a frozen language embedding space. This shared space additionally supports similarity-based retrieval, provides MaxCosine directly at inference time as an intrinsic reliability-ranking signal, and enables the representation to be evaluated along semantic dimensions that were not used as training labels.

The conventional decoding results address the first study question by showing that the language-aligned representation retains the fixed-intent information required for category discrimination. Top-$k$ and macro-F1 performance were closely matched between the two systems, while balanced 2v2 accuracy remained clearly above chance, with subject-level means of approximately 0.69--0.73. These findings establish an important performance anchor: moving from a conventional class output to a frozen language embedding space preserved comparable decoding performance while providing a representation that can support additional similarity-based analyses.

The second study question concerns whether the similarity structure itself carries useful information about prediction reliability. Across all five subjects, validation-derived MaxCosine thresholds transferred closely from validation to held-out test data, and retaining the highest-scoring trials increased both Top-1 and Top-5 accuracy for every participant. At the approximately 20\% operating point, Top-1 accuracy increased from 8.3--10.0\% at full coverage to 15.6--19.2\%, while Top-5 accuracy increased from 29.2--32.7\% to 42.9--50.5\%. Fold-wise AUROC values also remained above chance across subjects. Because MaxCosine is obtained directly from the EEG--prototype similarities used for decoding, these results show that the same retrieval score can provide an intrinsic reliability-ranking signal without requiring an additional confidence model.

The third study question examines whether the learned representation contains semantic organization beyond the original 39-category decision. Mean balanced accuracy exceeded the 50\% chance reference across all five evaluation-time semantic axes, with the clearest correspondence for Practical--Emotional, Goal-directed--Leisure, and Physical--Cognitive. The SGP provides a complementary qualitative view by showing that a held-out EEG embedding can be compared directly with alternative candidate sentences that were not used during EEG-encoder training. Together, these analyses indicate that the frozen language space provides structure beyond a conventional class output and may support ranked retrieval, candidate-set expansion, and future user-in-the-loop confirmation in communication-oriented BCI systems~\cite{edelman2025noninvasive, DalSeno2010Utility, Ma2023BCIUtility, song2022asyncmi}.

Several considerations define the scope of these findings. Although both systems use the same EEG data, preprocessing, cross-validation splits, and held-out test trials, they employ task-specific architectures and different checkpoint-selection criteria. Their comparison is therefore system-level and does not isolate the effect of contrastive learning under an identical backbone. Moreover, MaxCosine is derived from the same similarity space used for category decoding and is retained as a raw score. The present analysis demonstrates reliability ranking but does not establish probabilistic calibration. All experiments were conducted within-subject on a single dataset of healthy participants. The results therefore establish subject-specific feasibility but do not yet determine robustness across sessions, recording conditions, participant populations, or long-term use~\cite{edelman2025noninvasive}. Although the model operates in a language embedding space, training supervision and the principal decoding metrics remain tied to a fixed set of 39 intents and specific textual anchors. The semantic-axis analysis extends evaluation to dimensions introduced only after training; however, the axes are manually defined and the reference assignments are derived from sentence embeddings within the same dataset. The present results therefore support semantic correspondence beyond the original category taxonomy for the evaluated dimensions. Establishing unrestricted open-vocabulary generalization will require systematic testing with broader paraphrase sets, alternative formulations, and independently constructed semantic targets. More broadly, imagined-speech EEG remains a low-SNR and highly variable paradigm, with substantial sensitivity to artifacts, limited dataset scale, and unresolved challenges in reproducibility and deployment \cite{lopezbernal2022review}. In addition, the selective decoding analysis is performed only offline, not in a full asynchronous control loop with an explicit idle state, online timing constraints, or interactive confirmation.

These limitations also define the next steps. A first priority is to improve representation quality through stronger preprocessing, better artifact control, and training strategies that make better use of the available signal structure. The Reading condition in Chisco may provide a useful supervisory signal for pre-training before adaptation to imagined speech, particularly when combined with EEG foundation models trained on larger and more diverse datasets. A second priority is to extend semantically aligned EEG representations beyond the fixed-intent setting studied here. A natural next step is to test progressively more flexible retrieval and generation pipelines, beginning with larger candidate sets and systematic paraphrase evaluation before moving toward open-vocabulary decoding. Progress will also depend on broader datasets, more subjects and sessions, and adaptation strategies that combine shared modeling with lightweight user-specific calibration. Future studies should also evaluate semantic alignment in interactive BCI settings, where MaxCosine or related similarity scores could be tested as online criteria for abstention, shortlist presentation, or confirmation-based control.

\section{Conclusion}
\label{sec:conclusion}

This work investigated whether imagined-speech EEG can be aligned with a frozen language embedding space without sacrificing conventional fixed-intent decoding performance. Across five subjects in the Chisco dataset, the proposed contrastive framework achieved Top-$k$, macro-F1, and balanced pairwise performance comparable to a task-optimized closed-set classifier while representing single-trial EEG directly in a structured linguistic space.

The resulting representation provided capabilities beyond the conventional class output. Validation-derived MaxCosine thresholds consistently identified subsets of held-out trials with higher Top-1 and Top-5 accuracy, demonstrating that the similarity produced during retrieval carries useful information for reliability ranking. In addition, EEG embeddings showed mean agreement above the 50\% chance reference across all five semantic axes introduced only at evaluation time, while the Semantic Gallery Probe qualitatively demonstrated candidate-set flexibility through direct comparison with alternative sentences not used during EEG-encoder training.

Taken together, these results position contrastive EEG--language alignment as a promising representation framework for imagined-speech decoding: one that retains comparable fixed-vocabulary decoding while supporting ranked retrieval, intrinsic reliability ranking, and evaluation of broader semantic structure. Extending these findings across subjects, sessions, datasets, and interactive BCI settings will be important for determining how far these representational advantages can translate into more flexible communication systems.


\section*{References}

\bibliographystyle{IEEEtran}
\bibliography{references}

\end{document}